\documentclass[runningheads]{llncs}
\usepackage{graphicx}
\usepackage{microtype}
\usepackage[ruled,linesnumbered]{algorithm2e}
\usepackage{epstopdf}
\usepackage[utf8]{inputenc}
\usepackage{multirow}
\usepackage{multicol}
\usepackage{makecell}
\usepackage{threeparttable}
\usepackage{CJKutf8}
\usepackage{xstring}
\usepackage{booktabs}
\usepackage{graphicx}
\usepackage{tabularx}
\usepackage{soul}
\usepackage{color}

\usepackage{ulem} 
\usepackage{subfigure} %需要使用的宏包

\usepackage[misc,geometry]{ifsym}
\begin{document}

\title{I-WAS: a Data Augmentation Method with GPT-2 for Simile Detection \thanks{Supported by the Key Research and Development Program of Zhejiang Province (No. 2022C01011)} }

% \author{First Author \inst{1}  (\Letter) \orcidID{0000-1111-2222-3333} \and

\author{Yongzhu Chang \inst{1}  (\Letter) \orcidID{0000-0001-8418-9899} \and
Rongsheng Zhang\inst{1}\orcidID{0009-0008-1248-2090} \and
Jiashu Pu\inst{1}\orcidID{0000-0002-0549-9563}}

% \authorrunning{Yongzhu. C. et al.}
\institute { Fuxi AI Lab, NetEase Inc., Hangzhou, China \\ \email{
 \{changyongzhu, zhangrongsheng, pujiashu\}@corp.netease.com}}

% \setkeys{GPT-2}{}

\maketitle
% \begin{abstract}
% The abstract should briefly summarize the contents of the paper in

% \keywords{First keyword  \and Second keyword \and Another keyword.}
% \end{abstract}

\begin{abstract}

  Simile detection is a valuable task for many natural language processing (NLP)-based applications, particularly in the field of literature. However, existing research on simile detection often relies on corpora that are limited in size and do not adequately represent the full range of simile forms. To address this issue, we propose a simile data augmentation method based on \textbf{W}ord replacement \textbf{A}nd \textbf{S}entence completion using the GPT-2 language model. Our iterative process called \textbf{I-WAS}, is designed to improve the quality of the augmented sentences. To better evaluate the performance of our method in real-world applications, we have compiled a corpus containing a more diverse set of simile forms for experimentation. Our experimental results demonstrate the effectiveness of our proposed data augmentation method for simile detection.

  
\end{abstract}

\begin{CJK}{UTF8}{gbsn}
%\begin{document}
% \iftaclpubformat
\section{Introduction}

Figurative language, particularly analogy, is a common feature of literature and poetry that can help to engage and inspire readers~\cite{pual}. An analogy is a type of figurative language that compares two different objects in order to make a description more vivid. An analogy can take the form of a metaphor or a simile. Unlike metaphors, similes explicitly use comparative words such as ``\textit{like}", ``\textit{as}", and ``\textit{than}" in English, or ``像" (Xiang), ``如同" (Ru Tong), and ``宛若" (Wan Ruo) in Chinese. Simile sentences also typically have both a TOPIC, which is the noun phrase that acts as the logical subject, and a VEHICLE, which is the logical object of the comparison and is typically a noun phrase.

The Simile Detection (SD) task is crucial for various applications, such as evaluating student essays or extracting striking sentences from literature. Previous research on SD can be broadly classified into two categories: (1) studies that focus on identifying the sentiment conveyed by a given simile sentence, such as irony, humor, or sarcasm;~\cite{Niculae2014BrighterTG,Qadir2015LearningTR,Qadir2016AutomaticallyII} and (2) studies that aim to determine whether a sentence is a simile or not \cite{Weiguang2008ComputationOC,Veale2012ACM,Miwa2016EndtoEndRE}. Our paper focuses on the second type of simile detection. Many previous studies \cite{Ren2020AHR,Zhang2019CombiningPT,Zeng_Song_Su_Xie_Song_Luo_2020,Song2021AKG} on simile detection have used datasets that are limited in scope, specifically, datasets in Chinese that only contain examples with the same comparative word (e.g. ``like") and in which the TOPIC always appears to the left of the comparative word and the VEHICLE always appears to the right. Nevertheless, in real-world situations, as shown in Table~\ref{sentence}, simile sentences can involve a variety of comparative words, and the position of the TOPIC and VEHICLE relative to each other is not fixed. Additionally, existing datasets \cite{Weiguang2008ComputationOC,Liu2018NeuralML,Song2021AKG,Chakrabarty2022FLUTEFL} for simile detection are often of limited size. To develop an industrial simile detector, a corpus containing abundant examples and diverse simile forms is necessary. However, building such a corpus can be time-consuming and labor-intensive.

\begin{table*}[t]
    \centering
    \small
    \setlength{\tabcolsep}{20mm}{
    \resizebox{0.8\textwidth}{!}{%
    \begin{tabular}{@{}l@{}}
    \toprule
     \textbf{\textsf{TOPIC}}:天气 (Weather) \textbf{\textsf{VEHICLE}}: 火炉 (furnace) \\ \midrule
     \setlength{\tabcolsep}{14mm}{
    \begin{tabular}[c]{@{}l@{}}zh: 八月的天气 (TOPIC) 就像 (CW) 是火炉 (VEHICLE) 一样烘烤着大地。\\ en: The weather in August is like a furnace baking the hot earth.\end{tabular}} \\ \midrule 
    \setlength{\tabcolsep}{14mm}{
    \begin{tabular}[c]{@{}l@{}}zh: 八月的天气 (TOPIC) 已经炎热的如同 (CW) 火炉 (VEHICLE) 一般\\ en: The weather in August has been as hot as a furnace.\end{tabular}}             \\ \midrule
    \setlength{\tabcolsep}{14mm}{
    \begin{tabular}[c]{@{}l@{}}zh: 到了八月份，外面如火炉 (VEHICLE) 般 (CW) 的天气 (TOPIC) \\ en: By August, the furnace like weather outside.\end{tabular}}           \\ \midrule
    \setlength{\tabcolsep}{14mm}{
    \begin{tabular}[c]{@{}l@{}}zh: 八月份的天气 (TOPIC) 跟火炉 (VEHICLE) 一样 (CW) \\ en: The weather in August is the same as a furnace\end{tabular}}                       \\ \midrule
    \setlength{\tabcolsep}{14mm}{
    \begin{tabular}[c]{@{}l@{}}zh: 这八月份的天气 (TOPIC) ，俨然 (CW) 是一个大火炉 (VEHICLE) 啊\\ en: The August weather, as if it is a big furnace\end{tabular}}                       \\ \midrule
    \setlength{\tabcolsep}{14mm}{
    \begin{tabular}[c]{@{}l@{}}zh: 八月的天气 (TOPIC) 宛如 (CW) 火炉 (VEHICLE) 烘烤着炎炎大地\\ en: The August weather is similar to a furnace baking the hot earth\end{tabular}} \\ \bottomrule
    \end{tabular}%
    }
    }
    \caption{The examples of simile sentences that contain different comparative words. The relative positions of TOPIC, VEHICLE, and comparative words are not fixed. The CW means the comparative words.}
    \label{sentence}
\end{table*}

To address the issue of limited data availability in simile detection, we propose a data augmentation method that leverages \textbf{W}ord replacement \textbf{A}nd \textbf{S}entence completion with the GPT-2 language model. Our approach, called \textbf{WAS} (word replacement and sentence completion), involves the following steps: (1) Word replacement: for a given simile sentence, we randomly select a comparative word from a predefined set \footnote{Comparative words collected from Chinese similes} and replace the original comparative word in the sentence. (2) Sentence completion: using the replaced comparative word and the context above it as a prompt, we feed the modified simile sentence into the GPT-2 model to generate additional augmented sentences. We then use a simile detection model trained on the original corpus to rank the augmented sentences. Additionally, we investigate an iterative version of this process, called \textbf{I-WAS}, which aims to improve the quality of the augmented sentences by training a simile detection model on a mix of the original corpus and the augmented sentences in subsequent iterations.

To accurately evaluate the performance of simile detection models, we have compiled a corpus of simile sentences as a test set for experimentation. This corpus is more diverse than existing datasets, with a total of $606$ simile sentences that cover $7$ different comparative words and a range of TOPIC-VEHICLE position relations. The data is collected from the internet \footnote{http://www.ruiwen.com/zuowen/biyuju/} and manually labeled. We conduct extensive experiments on both the \cite{Liu2018NeuralML} dataset and our newly collected test set. The results of these experiments demonstrate the effectiveness of our proposed data augmentation method for simile detection.

Our contributions can be summarized as follows.

\begin{itemize}
	
    \item We propose a data augmentation method called \textbf{WAS} (\textbf{W}ord replacement \textbf{A}nd \textbf{S}entence completion) for generating additional simile sentences, and an iterative version of this process (\textbf{I-WAS}) to improve the quality of the augmented sentences.
    
    \item We have compiled a corpus of diverse simile forms for evaluating simile detection models, which will be made available upon publication.
    
    \item We conduct thorough experiments to demonstrate the effectiveness of our proposed data augmentation method for simile detection in real-world scenarios.
    
\end{itemize}

\section{Related work}

\subsection{Simile Detection}

Previous research on simile detection has focused on two main areas. The first area involves detecting the sentiment conveyed by a simile, such as irony, humor, or sarcasm \cite{Niculae2014BrighterTG,Qadir2015LearningTR,Qadir2016AutomaticallyII}. In this line of research, researchers have used rule-based, feature-based, and neural-based methods to identify the sentiment underlying simile sentences \cite{Qadir2015LearningTR,Qadir2016AutomaticallyII,Hao2010AnIF,Manjusha2018ConvolutionalNN}. For example, \cite{Chen2021JointlyIR} developed a classification module with a gate mechanism and a multi-task learning framework to capture rhetorical representation. The second area of research involves detecting whether a sentence is a simile or not \cite{Weiguang2008ComputationOC,Veale2012ACM,Miwa2016EndtoEndRE}. Feature-based approaches have been used for this purpose, with early work focusing on datasets containing only similes with the comparative word ``like" \cite{Liu2018NeuralML}. More recent studies have employed techniques such as bidirectional LSTM networks with attention mechanisms \cite{Guo2018AttentionBasedBN} and part-of-speech tags \cite{Zhang2019CombiningPT,Zeng2020NeuralSR} to improve the accuracy of simile detection.

In this paper, we focus on the task of determining whether a sentence is a simile or not. As illustrated in Table~\ref{sentence}, simile expressions come in many different forms and can be quite varied. While it is possible to extend existing methods to handle more complex and diverse simile sentences, to the best of our knowledge, the datasets provided by previous research \cite{Weiguang2008ComputationOC,Liu2018NeuralML,Song2021AKG} are not sufficient for evaluating the performance of simile detection models. Furthermore, there is currently no large-scale, annotated corpus for simile detection that is suitable for supporting data-driven approaches.

\subsection{Text Data Augmentation}

Variational autoencoders (VAEs) \cite{Kingma2014AutoEncodingVB}, generative adversarial networks (GANs) \cite{Goodfellow2014GenerativeAN}, and pre-trained language (PTL) generation models \cite{AnabyTavor2020DoNH,Kumar2020DataAU,Li2020ConditionalAF,Li2021TargetAwareDA} are commonly used for data augmentation in sentence-level sentiment analysis and text mining tasks \cite{Gupta2019DataAF,Hu2017TowardCG}. These methods typically involve encoding the input text into a latent semantic space and then generating new text from this representation. Nevertheless, these approaches are often not capable of generating high-quality text. Back translation, where text is translated from one language (e.g., English) into another (e.g., French) and then back again, has also been used for data augmentation \cite{Edunov2018UnderstandingBA}. However, this method is less controllable in terms of maintaining context consistency. Another class of data augmentation methods involves replacing words in the input text with synonyms or other related words. For example, \cite{Zhang2015CharacterlevelCN} and \cite{Wang2015ThatsSA} used WordNet \cite{Miller1995WordNetAL} and Word2Vec \cite{Mikolov2013DistributedRO} to identify words that could be replaced, while \cite{Wu2019ConditionalBC} and \cite{Kobayashi2018ContextualAD} used pre-trained models to predict the appropriate replacement words. \cite{Karimi2021AEDAAE} proposed inserting random punctuation marks into the original text to improve performance on text classification tasks. Recently, pre-trained language models have become popular due to their strong performance. \cite{Hou2018SequencetoSequenceDA} proposed a seq2seq data augmentation model for language understanding in task-based dialogue systems, and similar approaches have been used in other fields as well \cite{Claveau2021LaGD,Papanikolaou2020DAREDA}. Meanwhile, in the field of aspect term extraction, many studies have also employed pre-trained models for data augmentation \cite{Kober2021DataAF,Liu2020TellMH,Li2020ConditionalAF}.

However, none of the aforementioned data augmentation methods have been applied to the simile detection task, as they tend to operate at the sentence level and may unintentionally alter the simile components of simile sentences during the augmentation process.

\section{Task and Methodology}

\subsection{Formulation of Simile Detection}

Suppose we have a training dataset of size $n$, denoted as $D_{train}$ = {$(x_i, y_i)_{i=1}^n$}, where $(x_i^T, x_i^C, x_i^V) \in x_i$ represents the indices of the TOPIC, comparative word, and VEHICLE in a simile sentence, respectively, and $y_i \in$ \{$0, 1$\} indicates whether the sentence is a simile or not. Our data augmentation method aims to generate a new sentence $\hat{x_i}$ using the context of the original sentence $x_i$ (specifically, the TOPIC) as a prompt while maintaining the same label as the original sentence. The generated data is then combined with the original data and used as input for training a classifier, with the corresponding labels serving as the output.

\subsection{Simile Sentences Augmentation}

Previous conditional data augmentation approaches, such as those proposed by \cite{Wu2019ConditionalBC,Kumar2020DataAU}, can generate samples for datasets with unclear targets. These approaches \cite{Ding2020DAGADA,Kumar2020DataAU,Schick2021GeneratingDW} typically involve converting the sentences into a regular format and then using pre-trained models to generate new sentences through fine-tuning. In contrast, our proposed method, called \textbf{I}terative \textbf{W}ord replacement \textbf{A}nd \textbf{S}entence completion (I-WAS), generates TOPIC-consistent and context-relevant data samples without fine-tuning. It is important to maintain label consistency between the original and generated samples in data augmentation tasks \cite{Kumar2020DataAU}. To ensure label consistency in our approach, we first train a base simile detection model on the original corpus, replacing comparative words randomly. Then, we apply GPT-2 to generate 10 data samples for each sentence using Top-K sampling. Finally, we select the final label-consistent augmented samples using the base simile detection model to predict all of the augmented samples. We apply our augmentation method to the dataset proposed by \cite{Liu2018NeuralML}. The steps involved in our method are as follows:

\begin{algorithm}[h]
    \caption{Training}
    \label{augmentation}
    \KwIn{Training dataset $D_{\rm train}$; Total number of cycle $M$; Model parameters $\theta_{\rm basic}$ of simile detection; Model parameter $\theta_{\rm GPT-2}$}
    \KwOut{$\theta$}
    $\hat D_{\rm train}$: the augmentation dataset \;
    % $L$: the length of the dataset $D_{\rm train}$ \;
    $P$: the dataset with simile probability \;
    $G$: augmented samples dataset generated from GPT-2 \;
    $D_0$: the dataset after word replacement \;
    \textbf{select}:select augmented sample with label consistent with $y_j$\;
    \textbf{train}: training the model with $\theta$ \;
    \textbf{wr}: Comparative word replace with predefined set in Table~\ref{fenbu} \;
    $\theta$ = $\theta_{\rm basic}$ \;
    \For{$i\leftarrow 1$ \KwTo $M$}
    {
        $D_0 \leftarrow$  \textbf{wr}($D_{\rm train}$) \;
        $G \leftarrow$ Model($\theta_{\rm GPT-2}$, $D_0$) \;
        $P \leftarrow$ Model($\theta$, $G$) \;
        $\hat D_{\rm train} \leftarrow$ \textbf{select}(P) \;
        $D_{\rm train} \leftarrow $ $D_{\rm train}$ $\cup$ $\hat D_{\rm train}$ \;
        $\theta \leftarrow $ \textbf{train}$(\theta, D_{\rm train})$ \;
   }
    return $\theta$

\end{algorithm} 

\textbf{Word Replacement (WR): } The original dataset only includes the comparative word ``like". To add diversity to the dataset, we replace ``like" with a randomly selected comparative word from a predefined set, such as ``seems to" and ``the same as". For example, the original sentence \textit{``The weather in August is \textbf{like} a furnace baking the hot earth"} becomes \textit{``The weather in August is \textbf{the same as} a furnace baking the hot earth"} after replacement.

\textbf{Sentence Completion (SC): } To generate candidate samples, we mask the context after the comparative word and use the resulting prompt as input for the GPT-2 model. For example, given the original sentence \textit{``The weather in August is same as a furnace baking the hot earth"} the input for GPT-2 would be \textit{``The weather in August is same as."}. We follow the same process to generate candidate samples where the VEHICLE is located to the left of the comparative word. GPT-2 generates candidate samples through sentence completion using an auto-regressive approach and the top-k sampling strategy. We set a maximum length of 50 for the generated text and a size of 10 for the candidate set.

To select a suitable sample from the candidate set with a consistent label, we follow the following procedure: first, we obtain the labels of the original sentences and the probabilities predicted by the simile detection model. Then, we select the label-consistent augmented samples by maximizing or minimizing the probability, depending on the label of the original sentence. For example, if the label is negative, we rank the probabilities of the augmented samples and choose the sentence with the minimum probability. If the minimum probability is greater than 0.5, we change the label of the augmented sample. If the label is positive for the original sentence, we choose the sample with the maximum probability. In this way, we can obtain an augmented sample for each sentence in the dataset, resulting in a new dataset $\hat D_{\rm train}$.

\textbf{Iterative Process: } In the iterative process, we begin by obtaining the model parameters $\theta$ from the previous iteration. If it is the first iteration, $\theta$ is initialized with $\theta_{\rm basic}$, which is trained on $D_{\rm train}$. We then replace the ``like" in the original sentences with new comparative words, such as ``seem". We use GPT-2 to generate the candidate set based on the replaced sentences and annotate the candidate samples with the model using the parameters $\theta$, resulting in an augmentation dataset $\hat D_{\rm train}$. Finally, we merge $D_{\rm train}$ and $\hat D_{\rm train}$ and perform a simile detection task on the combined dataset. The details of the algorithm are shown in Algorithm~\ref{augmentation}.

\section{Experiments Setup}

\subsection{Diverse Test Set}

In real-world scenarios, simile sentences often involve a wide range of comparative words, as shown in Table~\ref{sentence}. To address this issue, we have collected a new test set from the Internet, referred to as the Diverse Test Set (\textbf{DTS}), which consists of 1k simile sentences and covers a variety of comparative words. However, the data collected from the Internet is likely to contain noise. To obtain a high-quality validation dataset, we hire 5 professional annotators on a crowdsourcing platform \footnote{https://fuxi.163.com/productDetail/17} to annotate the collected data. The true label for a sentence is set to 1 (is) or 0 (not), where 1 indicates that the sentence is a simile. We utilize a voting method to determine the labels for the final sentences. After filtering the data, we select $606$ simile sentences covering $7$ different comparative words. The number of simile sentences corresponding to different comparative words is shown in Table~\ref{fenbu}.

Meanwhile, to improve the quality of the dataset presented in this paper, we conduct a re-labeling process for the selected similes. Accordingly, we apply a set of $3$ criteria to evaluate the data: (1) Creativity (C), which refers to the level of originality or novelty of the sentence; (2) Relevance (R), which measures the extent to which the VEHICLE in the sentence is related to the TOPIC; and (3) Fluency (F), which evaluates the smoothness and clarity of the sentence. These criteria are rated on a scale from $1$ (not at all) to $5$ (very), and each sentence is evaluated by three students. A total of $20$, $15$, and $10$ students are used to rate the Creativity, Relevance, and Fluency of the similes, respectively. More information can be found in Table~\ref{rule_biaozhu}.

\begin{table*}[h]
    \centering
    % \small
    \resizebox{\textwidth}{!}{%
    \begin{tabular}{c|ccccccc}
    \toprule
    comparative words & 像(like) & 宛如(similar to) & 好似(seem) & 仿佛(as if) & 如同(as...as) & 跟…一样(same as) & 好比(just like) \\ \midrule
    count             & 333     & 74         & 72       & 56        & 38          & 27            & 6             \\ \bottomrule
    \end{tabular}%
    }
    \caption{The number of simile sentences with different comparison words in the DTS.}
    \label{fenbu}
\end{table*}

\begin{table}[h]
\centering
    \resizebox{0.5\textwidth}{!}{%
    \begin{tabular}{c|c|c|c}
        \toprule
            & Num & Average & $\alpha$ \\ \midrule
            Creativity (C) & 20 & 4.4 & 0.86 \\
            Relevance (R) & 15 & 3.9 & 0.75 \\
            Fluency (F) & 10 & 4.1 & 0.81 \\ \bottomrule
        \end{tabular}%
        }
    \caption{Num means the number of workers employed for each labeling task and $\alpha$ denotes the score of the reliability proposed by \cite{Krippendorff2011ComputingKA} to measure the quality of results provided by annotators.}
    \label{rule_biaozhu}
\end{table}

\begin{table*}[h]
\centering
\setlength{\tabcolsep}{14mm}{
    \resizebox{\textwidth}{!}{%
    \begin{tabular}{c|c|c}
        \toprule
            & BTS & DTS \\ \midrule
            Sentences & 2262 & 606 \\
            Simile sentences & 987 & 606 \\
            Non-simile sentences & 1275 & 0 \\
            Tokens & 66k & 16k \\
            Comparative Words & 1 & 7 \\ \bottomrule
        \end{tabular}%
        }
        }
    \caption{The statistical analysis of BTS and DTS. It is to verify the results of the experiments}
    \label{staticanalysis}
\end{table*}

\subsection{Dataset} \label{mysection}

In this paper, we apply the dataset proposed by \cite{Liu2018NeuralML} as our training and test sets. The test set is referred to as the \textbf{B}iased \textbf{T}est \textbf{S}et (\textbf{BTS}). The training set consists of $11337$ examples, including $5088$ simile sentences and $6249$ non-simile sentences. We follow \cite{Liu2018NeuralML} to divide the dataset into $7263$ training examples and $2263$ testing examples. However, the BTS contains only the comparative word ``like". Thus we provide a \textbf{D}iverse \textbf{T}est \textbf{S}et (\textbf{DTS}) in this paper. In addition to the comparative word ``like", our dataset includes 7 types of comparative words. The statistical analysis of the BTS and DTS datasets is shown in Table \ref{staticanalysis}. It is worth noting that the BTS from \cite{Liu2018NeuralML} contains both simile and non-simile sentences, while the DTS collected in this study only includes simile sentences. Finally, we apply two test sets to evaluate the performance of simile detection models.

\subsection{Baselines}

We compare the proposed augmentation approaches with the following baselines:

\textbf{EDA \cite{Wei2019EDAED}: } A simple way to augment data consists of four operations: synonym replacement, random swap, random insertion, and random deletion.

\textbf{BT \cite{2018Fast}: } This method involves translating a Chinese sentence into English and then translating the resulting English sentence back into Chinese.

\textbf{MLM \cite{Kobayashi2018ContextualAD,Kumar2020DataAU}: } The sentence is randomly masked and then augmented with a pre-trained model by contextual word vectors to find top-n similar words.

To assess the effectiveness of various data augmentation techniques, we apply the Bert-base model \cite{Devlin2019BERTPO}\footnote{https://huggingface.co/bert-base-chinese} as the base classifier. The Bert-base model is a $12$-layer transformer with $768$ hidden units and $12$ attention heads and has a total of 110M parameters. We conduct ablation tests to compare the performance of the data augmentation methods proposed in this paper. The specific methods and corresponding ablation tests are listed below:

\textbf{I-WAS: } An iterative data augmentation method with word replacement and sentence completion. More augmented data can be obtained during the iterative process. For example, I-$\rm WAS_1$ indicates the first iteration.

\textbf{I-WAS w/o wr: } An augmentation approach to sentence completion with a static comparative word such as ``like".

\textbf{I-WAS w/o sc: } A method without sentence completion that randomly replaces ``like" in simile sentences with different comparative words.

It is important to note that for each method of data augmentation, we only augment a single sample for each sentence in the original training dataset, $D_{\rm train}$. Additionally, the augmented dataset, $\hat D_{\rm train}$, is of size $I$-$1$ times that of the $D_{\rm train}$ dataset in the I-WAS method. This means that the size of the augmented dataset is determined by the number of augmentation iterations, $I$.

\begin{table*}[h]
    \centering
    % \resizebox{0.45\textwidth}{!}{%
    \setlength{\tabcolsep}{6mm}{
    \begin{tabular}{l|cc|c}
        \toprule
                        & \multicolumn{2}{c|}{BTS}   & DTS     \\ \midrule
        Methods                & $F_1$    &Accuracy      & Accuracy\\ \midrule
        Bert-base            & 87.54       & 88.61      & 66.70 \\
        + EDA \cite{Wei2019EDAED}              & 87.59       & 88.52     & 69.11 \\
        + BT  \cite{2018Fast}            & 87.35    & 88.15     & 72.48 \\
        + MLM \cite{Kumar2020DataAU}         & 87.79       & 89.08  & 70.20 \\ \midrule
        + I-$\rm WAS_1$              & 87.63       & 88.53    &\textbf{73.73} \\ 
        + I-$\rm WAS_1$ w/o sc         & \textbf{88.43} & \textbf{89.55}      & 70.30 \\
        + I-$\rm WAS_1$ w/o wr        &87.03        & 88.05     & 68.84\\
        \bottomrule
        \end{tabular}%
        }
        \caption{Comparison among different data augmentation methods on the biased test set and diversity test set. Bold indicates the best result, and the score means the average result of 5 runs with different seeds. The I-$\rm WAS_1$ means the first iteration of I-$\rm WAS$.}
    \label{tab:result}
\end{table*}

\subsection{Settings}

\textbf{Training details: } In all the experiments presented in this paper, we apply the same parameter settings for the simile detection task. We initialize the weights with the model that achieved the best performance in the previous iteration. The maximum sequence length is set to $256$, and the AdamW \cite{Loshchilov2019DecoupledWD} optimizer is used with a learning rate of 2e-4. The batch size is $128$, and we utilize an early stop of $3$ to prevent overfitting. For the sentence completion stage, we apply the GPT-2 \cite{Radford2019LanguageMA} model, which is trained on a large corpus of novels and common crawl and has $5$ billion parameters. The model uses a transformer architecture and consists of a $45$-layer decoder with $32$ attention heads in each layer. The embedding size is $3027$ and the feed-forward filter size is $3072$ * $4$. To generate diverse similes, we obtain $10$ samples using the GPT-2 model with the top-k sampling algorithm. All experiments are repeated using $5$ different random seeds ($16$, $32$, $64$, $128$, and $256$) and the average scores are reported.

\textbf{Metrics: } To facilitate comparison with previous work, we apply the same evaluation metrics as \cite{Liu2018NeuralML}, which are accuracy and the harmonic mean $F_1$ score. Since the DTS includes only simile samples, we focus on the evaluation metric of accuracy. 

\section{Results and Analysis}
\subsection{Results}

The experimental results on BTS and DTS are shown in Table~\ref{tab:result}, and the following conclusions can be drawn from the results.

\textbf{The effect of the test set: } We can observe that the accuracy of different data augmentation methods on the DTS is at least $15$ points lower than on the BTS. For instance, the accuracy of the I-$\rm WAS_1$ method on the DTS is $73.73$, while it is $88.53$ on the BTS. This difference is likely due to the fact that the simile sentences in the BTS only contain the comparative word ``like", while the DTS includes more diverse and realistic sentence expressions with various comparative words. These results suggest that the BTS proposed by \cite{Liu2018NeuralML} may not be representative of simile detection in real-world scenarios.

\textbf{I-WAS results: } Our proposed I-WAS data augmentation method achieves the highest accuracy score on the DTS compared to the other evaluated baselines (EDA, BT, MLM). Specifically, the I-$\rm WAS_1$ method increases the accuracy score from $66.70$ (using the Bert-base model) to $73.73$, representing an absolute improvement of $7.03$. Additionally, We also observe no significant differences in the performance of the various augmentation methods on the biased test set (BTS). These results demonstrate that the I-WAS method is particularly effective on realistic test sets.

\textbf{Ablation test results: } The results of the ablation test indicate that each of the \textit{\textbf{sc}}, \textit{\textbf{wr}}, and iterative process steps in the I-WAS method significantly improves the performance of simile detection on the DTS. Specifically, the gains achieved by the \textit{\textbf{sc}}, \textit{\textbf{wr}}, and iterative process steps were $3.43$ (from $70.30$ without \textit{\textbf{sc}} to $73.73$ with I-WAS), $4.89$ (from $68.84$ without \textit{\textbf{wr}} to $73.73$ with I-$\rm WAS_1$), respectively. Similar to the baseline methods, the experiments in the ablation test do not show significant improvements in performance on the BTS.

\begin{figure*}[h]

    \centering
    \subfigure[The Number of Iterations for I-WAS]{\includegraphics[width=0.4\linewidth]{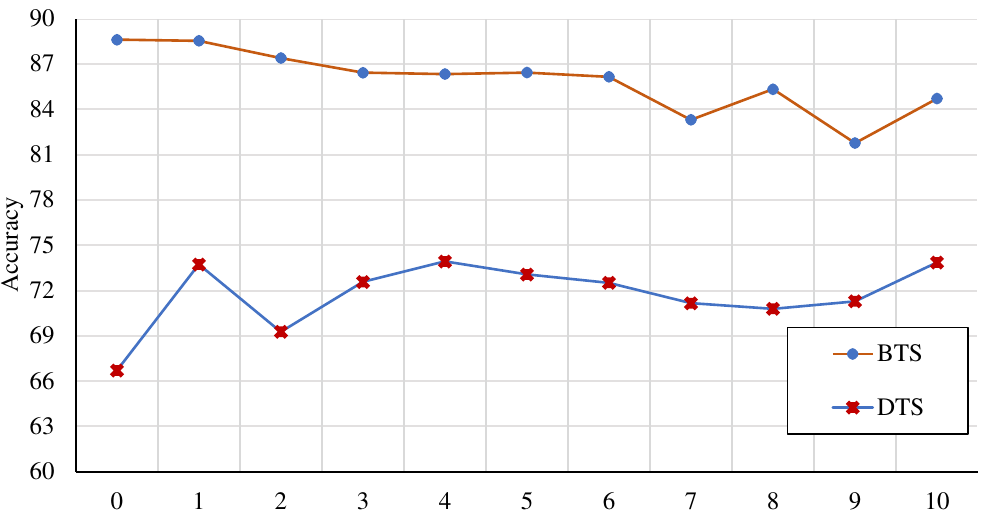}
    \setlength{\abovecaptionskip}{0.6cm}
    \label{fig:a}}
    \subfigure[Different fractions of the augmented samples]{\includegraphics[width=0.4\linewidth]{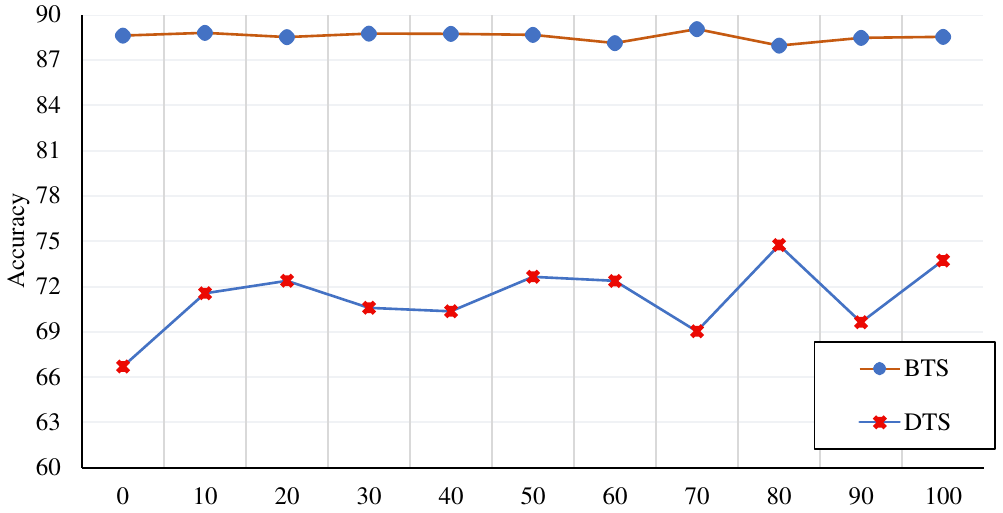} 
    \label{fig:b}}
    \caption{Performance of the model trained on different conditions. Figure (a) shows the model performance under various numbers of iterations, and the training data is twice as large as the original. Meanwhile, the pre-trained model is initialized with the optimal results of the previous iteration. Figure (b) indicates the impact of the number of augmented samples on the model performance. These fractions include \{10, 20, 30, 40, 50, 60, 70, 80, 90, 100\}\%. All the scores are the average of 5 runs with a different random seed. The y-axis shows the accuracy of the model.}
    \label{fig:iter}
    
\end{figure*}

\begin{table*}[h]
    \centering
    \setlength{\abovecaptionskip}{0.5cm}
    \resizebox{0.95\textwidth}{!}{%
    \begin{tabular}{c|c|c}
    \hline
    Source & Method  & Target                                                 \\ \hline
    \multirow{4}{*}{\begin{tabular}[c]{@{}c@{}} \\
    \\
    \\
    父母的爱就像天上的白云，永远不会减少。\\ Parents' love is like the white clouds in the sky, never diminishing.\end{tabular}} &
      EDA &
      \begin{tabular}[c]{@{}c@{}}父母的爱就\sout{像}天上的\sout{白云}，永远不会\sout{减少}。\\ Parents' love is \sout{like} the \sout{white clouds} in the sky, never \sout{diminishing}\end{tabular} \\ \cline{2-3} 
           & MLM     & \begin{tabular}[c]{@{}c@{}}父母的脸就像天上的白云，永远不会减少。\\ Parents' face is like the white clouds in the sky, never diminishing.\end{tabular} \\ \cline{2-3} 
           & BT      & \begin{tabular}[c]{@{}c@{}}父母的爱就像天上的白云，永远不会减少。\\ Parents' love is like the white clouds in the sky, never diminishing.\end{tabular}          \\ \cline{2-3} 
           & I-$\rm WAS_1$ & \begin{tabular}[c]{@{}c@{}}父母的爱就好似这颗大树，永远不会枯萎。\\ Parents' love is the same as the tree, never wither.\end{tabular}                  \\ \hline
    \multirow{4}{*}{\begin{tabular}[c]{@{}c@{}}
    \\
    \\
    书就像打开知识大门的金钥匙。\\ The Books are like the golden key to opening the door to knowledge.\end{tabular}} &
      EDA &
      \begin{tabular}[c]{@{}c@{}}打开就像书知识大门的金钥匙。\\ The opening is like the golden key to the door of book knowledge.\end{tabular} \\ \cline{2-3} 
           & MLM     & \begin{tabular}[c]{@{}c@{}}书就像打开知考大门的金钥锁。\\ The book is like the golden key lock to opening the door to the knowledge.\end{tabular}          \\ \cline{2-3} 
           & BT      & \begin{tabular}[c]{@{}c@{}}这本书就像将金钥匙打开到门口。\\ This book is like opening the golden key to the door.\end{tabular}                              \\ \cline{2-3} 
           & I-$\rm WAS_1$ & \begin{tabular}[c]{@{}c@{}}书就仿佛一座山。\\ The book is like a mountain.\end{tabular}                                                              \\ \hline
    \multirow{4}{*}{\begin{tabular}[c]{@{}c@{}}
    \\
    \\
    我大颗的泪水像细珠一样滚动着。\\ My large tears rolled like fine beads.\end{tabular}} &
      EDA &
      \begin{tabular}[c]{@{}c@{}}我大颗的\sout{泪水}像细珠一样滚动着。\\ My large \sout{tears} rolled like fine beads.\end{tabular} \\ \cline{2-3} 
           & MLM     & \begin{tabular}[c]{@{}c@{}}太大颗的泪都像细珠一样跳动着。\\ Too large tears rolled like fine beads.\end{tabular}                                            \\ \cline{2-3} 
           & BT      & \begin{tabular}[c]{@{}c@{}}我的眼泪像珠子一样滚动。\\ My tears rolled like beads.\end{tabular}                                                           \\ \cline{2-3} 
           & I-$\rm WAS_1$ & \begin{tabular}[c]{@{}c@{}}我大颗的泪水像断线的珠子一样不要钱的往下掉。\\ My big tears fell like broken beads.\end{tabular}                                        \\ \hline
    \end{tabular}%
	}
    \caption{Examples of augmented simile sentences of different methods.}
    \label{tab:sample}
\end{table*}

\subsection{Analysis} \label{fenxi}

\textbf{Number of Iterative Processes: } Figure \ref{fig:a} shows the influence of the number of iterations on the performance of the model. We set the number of iterations to $10$ and initialized the pre-trained model with the optimal model parameters from the previous iteration. As the number of iterations increases, the performance of the model on the BTS gradually decreases, while the performance on the DTS improves. This can be attributed to the data distribution drift \cite{Lesort2021UnderstandingCL}. During the continuous iterative learning process, the increasing number of augmented samples leads to a gradual shift in the distribution of the training data toward the DTS. As shown in Table~\ref{tab:result}, at the first iteration, the accuracy on the DTS improved by 7\% while the accuracy on the BTS decreased by 0.1\%, indicating the effectiveness of the proposed method in this paper.

\textbf{Trend on Training Set Sizes: } Figure \ref{fig:b} shows the performance of the model on different fractions of the augmented samples, where the largest fraction is twice the size of the original training data. These fractions include \{$10$, $20$, $30$, $40$, $50$, $60$, $70$, $80$, $90$, $100$\}. It is worth noting that adding different proportions of augmented samples has varying levels of impact on the model's performance.

As we can see from the figure, the model's performance on the DTS fluctuates more dramatically than that on the BTS when different proportions of augmented samples are inserted. This is because the distribution of the original training data is similar to that of the BTS, and the various fractions of augmented samples with noise can differently impact the original distribution of the training data. When the number of augmented samples is smaller, the influence of noisy data becomes more significant. For example, if there are $10$ enhanced samples and $5$ mislabeled samples, the noise data accounts for 50\%; when the number of enhanced samples increases to $100$ and the number of mislabeled samples increases to $20$, the percentage of noise data is reduced to 20\%.

\subsection{Augmented sample}

In this section, to evaluate the quality of the data generated by the different data augmentation approaches, we provide some examples in Table~\ref{tab:sample}. We can see that, with the exception of the EDA approach, all other approaches can maintain sentence completeness, coherence between the TOPIC and VEHICLE, and label consistency. Meanwhile, the effectiveness of the BT-based sentence expansion approach depends on the complexity of the original sentence. Simple sentences are more difficult to augment using this method, as shown in the first sample in Table~\ref{tab:sample}. On the other hand, complex sentences are easier to augment using this method. The MLM-based approach can change the fluency of sentences and potentially replace the TOPIC. However, the samples generated using the approach proposed in this paper increase the diversity of the VEHICLE without changing the TOPIC of the sentences, while maintaining label consistency and content fluency.

\section{Discussion}

Comparing the results of the BTS and DTS obtained in our experiments, we can see a significant gap, particularly in the performance of the Bert-base model. We believe that this gap may be due to the inconsistency between the training and test sets. As mentioned in Section \ref{mysection}, the training set used in our experiments is obtained from \cite{Liu2018NeuralML} and is consistent with the BTS. It only includes one simile pattern. However, the DTS includes several types of simile patterns, which means that a model trained on the BTS, such as the Bert-base model, may perform well on a particular pattern but not as well on others. This leads to a decrease in performance on the DTS. Additionally, the more simile patterns included in the test set, the more pronounced the decrease in model performance becomes.

In Figure~\ref{fig:a}, we observe a gradual decrease in the performance of the model on the BTS during the iteration process. \cite{Lesort2021UnderstandingCL} suggests that catastrophic forgetting may be caused by changes in the data distribution or changes in the learning criterion, which are referred to as drifts. \cite{Tarvainen2017MeanTA} propose Mean Teacher to overcome the problem that the targets change only once per epoch, and Temporal Ensembling becomes unwieldy when learning large datasets. In general, the data distribution changes over time in continual learning (or iterative learning). The decrease in performance shown in Figure~\ref{fig:a} is likely due to the changing data distribution of the original training set as the number of augmented samples increases. As the augmented samples include more patterns of similes, the model's performance in detecting these patterns improves. However, the performance on pattern-specific similes, such as those in the BTS, decreases. The problem of catastrophic forgetting in the process of continuous learning is an area for future work.

\section{Conclusion and Future work}

In this paper, we introduce a data augmentation method called WAS, which uses GPT-2 to generate figurative language for simile detection. Unlike existing augmentation methods that are unaware of the target information, WAS is able to generate content-relevant and label-compatible sentences through word replacement and sentence completion. Our experimental results show that WAS performs the best on the diverse simile test set, demonstrating the effectiveness of this generative augmentation method using GPT-2. Recent studies have shown that it is more effective to incorporate constraints during the generation process through cooperative language model generation. When using a classifier to filter the generated examples, it is important to consider the impact on the performance of the base simile detection model. Future research could focus on improving the performance of the base simile detection model and exploring related topics.

\end{CJK}

\section*{Acknowledgements}
This work is supported by the Key Research and Development Program of Zhejiang Province (No. 2022C01011). We would like to thank the anonymous reviewers for their excellent feedback. We are very grateful for the professional markers provided by NetEase Crowdsourcing.

\end{document}